\documentclass{article} 
\usepackage{lineno}
\usepackage{iclr2026_conference,times}
\usepackage{hyperref}
\usepackage{url}
\usepackage{graphicx} 
\usepackage{placeins}
\usepackage{caption}


\renewcommand{\arraystretch}{1.25}

\usepackage{amsmath,amsfonts,bm}

\def\eqref#1{equation~\ref{#1}}

\def\1{\bm{1}}

\DeclareMathAlphabet{\mathsfit}{\encodingdefault}{\sfdefault}{m}{sl}
\SetMathAlphabet{\mathsfit}{bold}{\encodingdefault}{\sfdefault}{bx}{n}

\title{RoPE-LIME: RoPE-Space Locality + Sparse-$K$ Sampling for Efficient LLM Attribution}

\author{Isaac Picov\thanks{Work done during an internship at DevRev.} \\
University of Toronto \\
Toronto, ON, Canada \\
\texttt{isaac.picov@mail.utoronto.ca} \\
\And
Ritesh Goru \\
DevRev \\
Austin, TX, USA \\
\texttt{ritesh.goru@devrev.ai}
}

\iclrfinalcopy 
\begin{document}

\maketitle
\lhead{} 

\begin{abstract}
Explaining closed-source Large Language Model (LLM) outputs is challenging because API access prevents gradient-based attribution, while perturbation methods are costly and noisy when they depend on regenerated text. We introduce \textbf{Rotary Positional Embedding Linear Local Interpretable Model-agnostic Explanations (RoPE-LIME)}, an open-source extension of gSMILE that decouples reasoning from explanation: given a fixed output from a closed model, a smaller open-source surrogate computes token-level attributions from probability-based objectives (negative log-likelihood and divergence targets) under input perturbations. RoPE-LIME incorporates (i) a locality kernel based on Relaxed Word Mover's Distance computed in \textbf{RoPE embedding space} for stable similarity under masking, and (ii) \textbf{Sparse-$K$} sampling, an efficient perturbation strategy that improves interaction coverage under limited budgets. Experiments on HotpotQA (sentence features) and a hand-labeled MMLU subset (word features) show that RoPE-LIME produces more informative attributions than leave-one-out sampling and improves over gSMILE while substantially reducing closed-model API calls.
\end{abstract}

\section{Introduction}
Large Language Models (LLMs) have become ubiquitous, yet their decision-making processes remain largely opaque. This interpretability gap is widened by the prevalence of closed-source models accessed solely via APIs, which precludes the use of gradient-based attribution methods. Consequently, perturbation-based approaches, which probe model behavior by modifying inputs, have become the primary means of explaining these black-box systems.

Recent works such as gSMILE \citep{Dehghani2025} have adapted the LIME framework \citep{Ribeiro2016} to the generative setting, employing weighted linear surrogates to attribute output tokens to input features. However, such methods face significant challenges: (1) \textbf{High Cost}: They require repeated queries to the closed-source model for every perturbation, making them prohibitively expensive for long contexts; (2) \textbf{Output Instability}: Relying on generated text rather than probability distributions introduces noise, as small perturbations can lead to vastly different generation trajectories; and (3) \textbf{Sampling Inefficiency}: Standard sampling strategies like Leave-One-Out (LOO) or random masking often fail to capture feature interactions or scale poorly with input length.

To address these limitations, we introduce RoPE-LIME, an open-source framework that decouples reasoning from explanation. The closed model is queried only once to generate the original response, while a smaller open-source surrogate computes probability-based attribution scores for that fixed output. This design drastically reduces cost and enables richer distributional metrics (e.g., KL divergence) instead of discrete text overlap. We further improve attribution quality with two innovations: (1) a locality kernel using Relaxed Word Mover’s Distance in RoPE space for more stable span comparison, and (2) Sparse-K sampling, a logarithmic-complexity perturbation strategy that efficiently explores the feature space.

\section{Related Work}
\label{sec:related_work}
\textbf{Perturbation-Based Attribution.}
Model-agnostic explanation methods like LIME \citep{Ribeiro2016} and SHAP \citep{Lundberg2017} estimate feature importance by training local surrogate models on perturbed inputs. While effective for classifiers, applying them to generative tasks is non-trivial due to the unbounded output space and the high cost of autoregressive generation. gSMILE \citep{Dehghani2025} extends LIME to LLMs by using Wasserstein distance metrics and generating text for perturbed prompts. Our work builds on this foundation but diverges by replacing the expensive closed-loop generation with an open-source surrogate, thereby enabling probability-based regression targets.

\textbf{Context Attribution \& Efficiency.}
Attributing model outputs to context is often formulated as a Leave-One-Out (LOO) problem. However, naive LOO is computationally expensive. Recent approaches like AttriBoT \citep{Liu2024} propose efficiency tricks for approximating LOO, such as caching and proxy models. Similarly, we leverage a proxy model (our open-source surrogate) but use it within a LIME-style regression framework rather than for direct LOO approximation. This allows us to capture feature interactions via our Sparse-K sampling, which scales logarithmically ($O(\log N)$) compared to the linear scaling of LOO.

\textbf{Geometric Distance Metrics.}
Defining "locality" for text perturbations requires a meaningful distance metric. Word Mover's Distance (WMD) \citep{Kusner2015} and its relaxed variant (RWMD) provide a robust measure of semantic distance between documents. We adapt RWMD to the era of modern Transformers by computing distances over Rotary Positional Embeddings (RoPE) \citep{Su2024}. Unlike absolute positional encodings, RoPE encodes relative positions via rotation, making our distance metric more robust to the index shifts caused by masking-based perturbations.

\section{Method}
\label{method}
\subsection{RWMD + RoPE}
We adopt Relaxed Word Mover's Distance (RWMD) to reduce complexity to $\mathcal{O}(n^2)$ while preserving optimal-transport structure \citep{Kusner2015}. Unlike prior WMD approaches using static embeddings, we compute RWMD over contextualized token embeddings for fine-grained attribution. To mitigate sensitivity to positional perturbations in absolute encodings, we employ Rotary Positional Embeddings (RoPE). Since RoPE aligns with the model's native attention mechanism, computing RWMD in this space provides a stable attribution proxy consistent with the model's inductive biases \citep{Su2024}.

\subsection{Feature Representations}

RWMD operates over sets of points with well-defined pairwise distances, requiring each feature to be represented as a single embedding regardless of its underlying token length \citep{Kusner2015}. Because we operate at the token level, features may correspond to arbitrary spans (e.g., sentences or paragraphs) and need not be uniform in length. Naively pooling token embeddings in Euclidean space fails to respect the rotational geometry induced by RoPE and introduces sensitivity to absolute index reassignment \citep{Su2024}. We therefore construct span-level representations directly in the polar domain, and use polar $L_2$ in place of standard $L_2$.

Let a feature correspond to a token span $S = \{x_1, \dots, x_m\}$, where each $x_i \in \mathbb{R}^d$ is a contextualized RoPE embedding. RoPE represents embeddings in $\mathbb{C}^{d/2}$ as rotations; for each dimension $k$,
\[
z_{ik} = x_{i,2k} + i\,x_{i,2k+1} = r_{ik} e^{i\theta_{ik}},
\]
where $r_{ik} = |z_{ik}|$ and $\theta_{ik} = \operatorname{atan2}(x_{i,2k+1}, x_{i,2k})$.
We aggregate span embeddings by averaging magnitudes and relative phases:
\[
\bar{r}_k = \frac{1}{m} \sum_{i=1}^m r_{ik}, 
\qquad
\bar{\theta}_k = \operatorname{Arg}\!\left(\sum_{i=1}^m e^{i\theta_{ik}}\right).
\]

Define polar $L_2$ as:
\[
d_{\text{polar}}(x,y) =
\sqrt{
\sum_k (\bar{r}_{xk}-\bar{r}_{yk})^2
+ \beta \sum_k (\bar{\theta}_{xk}-\bar{\theta}_{yk})^2
},
\]
with $\beta = 1$ for simplicity \citep{Su2024}. This construction preserves the relative phase structure induced by RoPE while remaining invariant to global index shifts.

\subsection{Sparse-K Sampling}
\label{sec:sparse_k}

Conventional perturbation-based sampling strategies either under-sample the feature space, leading to poor attribution quality, as observed with leave-one-out (LOO) sampling or scale impractically, as in powerset-based sampling. Common heuristics (e.g., $\sim$10 perturbations per feature) quickly exceed feasible computational budgets for realistic feature counts \citep{Peduzzi1996}.

We therefore propose Sparse-K sampling, a perturbation strategy that achieves strong empirical performance while scaling logarithmically with the number of features by exploiting contextual dependence among natural-language features. We define the perturbation budget as:
\[
\label{eq:sparse-k-budget}
N = c \cdot \log K
\]
where $N$ is the number of samples used to fit the regression model and, $c,K$ are hyperparameters chosen as functions of $M$, the number of input features. These choices balance estimation stability and computational cost; optimal configurations are determined empirically in \S\ref{sec:hotpot} and a full set of configurations can be found in Appendix~\ref{sec:sparsek_sweep}.

\subsection{RoPE-LIME Pipeline}

Our pipeline follows the LIME framework of \citet{Ribeiro2016} and incorporates design elements from gSMILE~\citep{Dehghani2025}. As in prior work, feature attributions are obtained by fitting a weighted linear surrogate model, with attribution scores derived from the magnitude of the learned coefficients. Sampling efficiency is achieved via Sparse-\(K\) perturbation, which substantially reduces the number of required model evaluations.

Unlike prior approaches, attribution is performed over text generated by a closed-source (or large open-source) model, while all explanation is computed using a smaller open-source surrogate. The surrogate attends over the fixed generated output and evaluates perturbed inputs, allowing attribution to be computed using probability-based regression targets without repeated closed-model calls. This decouples reasoning capture from attribution computation and enables scalable explanation.

All attribution is performed using a smaller open-source model $f_{\mathrm{S}}(\cdot):\mathbb{R}^d\mapsto \mathbb{R}^T$. $f_{\mathrm{S}}(\cdot)$ attends over the output text $y$, generating logits that indicate token likelihood with respect to masked input. This process allows us to use our small model as a surrogate to our reasoning model and decouples reasoning capture from attribution computation. From $\mathcal{F}$, we apply Sparse-K sampling to obtain $N$ perturbations and construct a sparse perturbation matrix $\mathbf{Z}\in\{0,1\}^{n\times N}$, with $\mathbf{z}_0=\mathbf{1}$ denoting the unmasked baseline.

Let $x$ denote the input prompt and $y = f_{\mathrm{L}}(x) = (y_1,\dots,y_T)$ the generated output from a large model $f_{\mathrm{L}}(\cdot)$ as text. We seek to attribute $y$ to a set of interpretable input features $\mathcal{F} = \{F_1,\dots,F_n\}$, where each $F_i$ corresponds to a contiguous span (e.g., sentence or paragraph).

For each perturbation $\mathbf{z}_j$, we define NLL to be Negative Log Likelihood and evaluate:
\[
\ell_j = f_{\mathrm{S}}(x\odot \mathbf{z}_j; y), \qquad
\mathcal{L}_j = \mathrm{NLL}(\ell_j),
\]

let $\mathbb{KL}$ be KL-divergence, then define regression targets relative to the baseline $\mathcal{L}_0$ as:
\[
y_j^{\mathrm{reg}} = \mathbb{KL}(\mathcal{L}_0 || \mathcal{L}_j)
\]

We compute our regression weights via RWMD which are used to enforce locality:
\[
d_j = \mathrm{RWMD}(x, x\odot \mathbf{z}_j), \qquad
w_j = \exp\!\left(-\frac{d_j^2}{\sigma^2}\right),
\]
Choosing $\sigma$ to be the median of our set $D = \{d_1,\dots,d_N\}$, from this we yield a diagonal weight matrix $\mathbf{W}=\mathrm{diag}(w_1,\dots,w_N)$.

Feature attributions are estimated via weighted least squares and normalized to obtain final attribution scores:
\[
\hat{\boldsymbol{\beta}}
= \arg\min_{\boldsymbol{\beta}}
\left\|
\mathbf{W}^{1/2}\big(\mathbf{y}^{\mathrm{reg}}-\mathbf{Z}\boldsymbol{\beta}\big)
\right\|_2^2,
\qquad
a_i = \frac{|\beta_i|}{\sum_j |\beta_j|}.
\]

Attribution scores $a_j$ quantify the contribution of each feature $F_j$; the intercept $\beta_0$ is omitted as it has no feature level interpretation..

\section{Results}
{\centering
\includegraphics[width=0.75\linewidth]{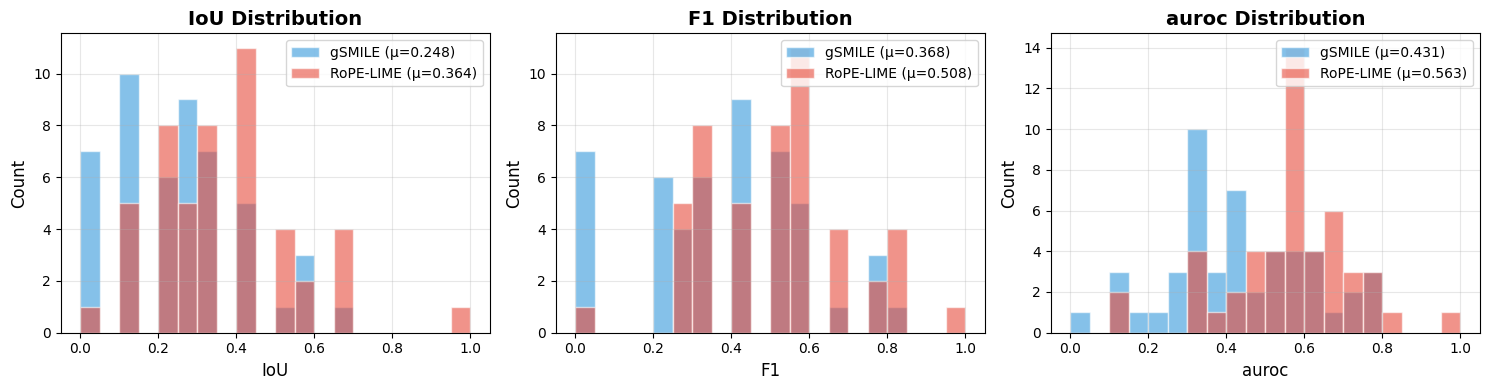}
\captionof{figure}{gSMILE (gpt-4o-mini) vs RoPE-LIME (Qwen-8B)}
\label{fig:iou_mmlu}}

\medskip
{\centering
\includegraphics[width=0.75\linewidth]{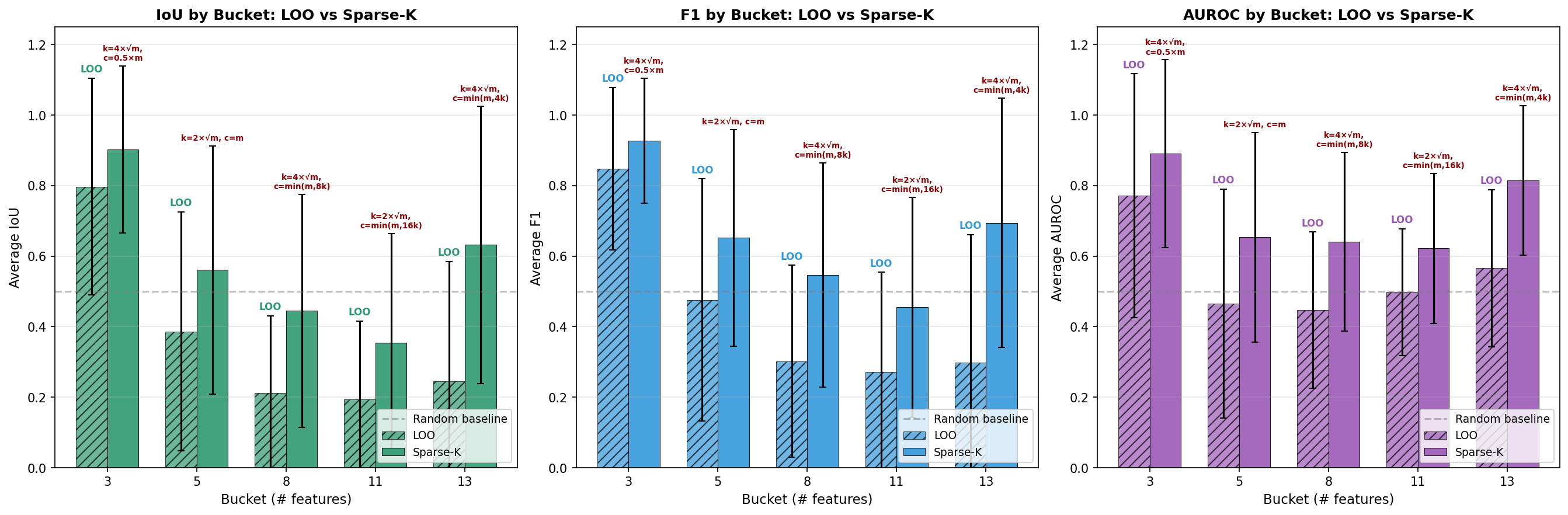}
\captionof{figure}{RoPE-LIME evaluated on longer contexts}
\label{fig:rope_lime_long}}
\vspace{3pt}

We evaluate our method on the MMLU and HotpotQA datasets, both of which are question-answering benchmarks designed to assess reasoning capabilities in large language models. For the purposes of this study, we leverage the structure of these datasets to evaluate attribution accuracy rather than answer correctness.

We report standard evaluation metrics, including F1, intersection-over-union (IoU), and area under the ROC curve (AUROC).

Closed-source inference is performed using gpt-4o-mini accessed via the ChatCompletions API, while open-source inference is conducted using a locally hosted Qwen-8B model running on an NVIDIA H200 GPU.

\subsection{Evaluation against gSMILE}
gSMILE evaluates its method on a hand-labeled subset of MMLU by treating each query word as a feature and measuring attribution quality against manually annotated influential words \citep{Dehghani2025}. We replicate this protocol using 50 queries from the miscellaneous category of MMLU, labeling the top five influential words per query, and evaluate our method under the same setup. Results are shown in Fig.~\ref{fig:iou_mmlu}, with full results reported in Table~\ref{tab:full-results} (Appendix).

To ensure a fair comparison, we do not apply Sparse-K sampling and instead use a fixed perturbation budget of 60 samples, matching the gSMILE protocol and isolating the effect of replacing closed-source reasoning. Under this evaluation, our method outperforms gSMILE across IoU, F1, and AUROC despite relying exclusively on smaller open-source models. While absolute performance remains modest, these results indicate that high-quality attribution on short, query-only inputs does not strictly require closed-source inference. At the same time, MMLU represents a constrained setting in which features are treated as independent words despite strong contextual dependencies, limiting potential gains as mentioned in \S\ref{sec:sparse_k}.

\subsection{Evaluation on HotpotQA}
\label{sec:hotpot}
Because full-scale evaluation using closed-source models is cost prohibitive, we evaluate our method on a labeled subset of HotpotQA independent of gSMILE. We select all examples containing exactly 10 retrieved documents, yielding 989 instances. Each sentence is treated as a feature, and examples are grouped into feature-count buckets of approximately 100 instances each, with ranges [2\textendash{}3],[4\textendash{}5],[6\textendash{}8],[9\textendash{}11],[12+] and corresponding sizes [96,100,99,99,24]. We normalize to 100 as our largest bucket (12+ features) is underrepresented in our dataset.

Using annotated supporting facts as ground truth, we evaluate Sparse-K sampling across multiple configurations within each bucket. Performance is measured using IoU, F1, and AUROC. Fig.~\ref{fig:rope_lime_long} reports the optimal configuration per feature-count bucket and compares Sparse-K against LOO sampling, with detailed results provided in Tables~\ref{tab:sparsek-configs} and~\ref{tab:hotpot-loo} (Appendix).

\section{Conclusion}
We present RoPE-LIME, an open-source attribution framework that extends gSMILE by decoupling reasoning from explanation and enabling probability-based attributions for closed-source LLM outputs. By combining RWMD over RoPE-based embeddings with Sparse-K sampling, our method produces stable attributions that scale to long contexts at substantially lower cost. Results on MMLU and HotpotQA demonstrate that faithful and practical LLM interpretability can be achieved without repeated closed-source inference.

\clearpage
\bibliography{iclr2026_conference}
\bibliographystyle{iclr2026_conference}
\nocite{*}

\clearpage

\appendix
\FloatBarrier

\section{Tables}

\begin{table}[h!]
\caption{gSMILE vs RoPE-LIME (50 Queries)}
\label{tab:full-results}
\begin{center}
\begin{tabular}{lcccc}
\multicolumn{5}{c}{\bf CLOSED-SOURCE} \\
\hline
Metric & Mean $\pm$ Std & Median [IQR] & Min & Max \\
\hline
IoU   & 0.248 $\pm$ 0.171 & 0.250 [0.222] & 0.000 & 0.667 \\
F1    & 0.368 $\pm$ 0.216 & 0.400 [0.300] & 0.000 & 0.800 \\
AUROC & 0.431 $\pm$ 0.182 & 0.400 [0.258] & 0.000 & 0.792 \\
\hline

\multicolumn{5}{c}{\bf OPEN-SOURCE} \\
\hline
Metric & Mean $\pm$ Std & Median [IQR] & Min & Max \\
\hline
IoU   & 0.364 $\pm$ 0.184 & 0.333 [0.229] & 0.000 & 1.000 \\
F1    & 0.508 $\pm$ 0.191 & 0.500 [0.267] & 0.000 & 1.000 \\
AUROC & 0.563 $\pm$ 0.159 & 0.583 [0.161] & 0.125 & 1.000 \\
\hline
\end{tabular}
\end{center}
\end{table}

\begin{table}[h!]
\caption{Sparse-$K$ configuration sweep on HotpotQA. Results reported as mean $\pm$ standard deviation.}
\label{tab:sparsek-configs}

\renewcommand{\arraystretch}{1.25}

\begin{center}
\begin{tabular}{cccccc}
\hline
$M$ & $N$ & $(k,c)$ &
IoU&
F1 &
AUROC \\
\hline
3  & 96  & $(4\sqrt{M},\,0.5M)$         & $0.903 \pm 0.237$ & $0.927 \pm 0.177$ & $0.891 \pm 0.266$ \\
5  & 100 & $(2\sqrt{M},\,M)$            & $0.561 \pm 0.351$ & $0.652 \pm 0.307$ & $0.653 \pm 0.297$ \\
8  & 99  & $(4\sqrt{M},\,\min(M,8k))$   & $0.445 \pm 0.330$ & $0.546 \pm 0.318$ & $0.641 \pm 0.253$ \\
11 & 99  & $(2\sqrt{M},\,\min(M,16k))$  & $0.355 \pm 0.310$ & $0.455 \pm 0.312$ & $0.622 \pm 0.213$ \\
13 & 24  & $(4\sqrt{M},\,\min(M,4k))$   & $0.632 \pm 0.393$ & $0.694 \pm 0.353$ & $0.815 \pm 0.212$ \\
\hline
\end{tabular}
\end{center}
\end{table}

\begin{table}[h!]
\caption{HotpotQA performance using leave-one-out (LOO) perturbations.
Values reported as mean $\pm$ standard deviation.}
\label{tab:hotpot-loo}

\renewcommand{\arraystretch}{1.25}

\begin{center}
\begin{tabular}{ccccc}
\hline
$M$ & $N$ &
Mean IoU &
Mean F1  &
Mean AUROC\\
\hline
3  & 96  & $0.797 \pm 0.307$ & $0.848 \pm 0.230$ & $0.772 \pm 0.345$ \\
5  & 100 & $0.386 \pm 0.339$ & $0.476 \pm 0.343$ & $0.466 \pm 0.324$ \\
8  & 99 & $0.212 \pm 0.219$ & $0.302 \pm 0.272$ & $0.447 \pm 0.221$ \\
11 & 99 & $0.193 \pm 0.223$ & $0.272 \pm 0.282$ & $0.498 \pm 0.180$ \\
13 & 24 & $0.246 \pm 0.339$ & $0.298 \pm 0.362$ & $0.566 \pm 0.222$ \\
\hline
\end{tabular}
\end{center}
\end{table}

\subsection{Sparse-K parameter sweep}
\label{sec:sparsek_sweep}

We performed a bounded parameter sweep over the Sparse-$K$ sampling hyperparameters $(k,c)$ to evaluate sensitivity to sampling density and budget allocation. The sweep is not intended to identify an optimal configuration, but to assess robustness across a range of practically reasonable settings.

We vary the sparsity parameter $k \in \{\sqrt{m},\,2\sqrt{m},\,4\sqrt{m}\}$, which controls the number of active features per perturbation. These values interpolate between very sparse masks ($\sqrt{m}$) and denser but still sublinear regimes ($4\sqrt{m}$), allowing us to evaluate the trade-off between locality and feature interaction coverage.

We vary the sample budget parameter
\[
c \in \{\min(m,16k),\, m,\, \min(m,8k),\, 0.5m,\, 0.25m,\, \min(m,4k)\},
\]
which determines the total number of perturbations $N$ as a function of feature count. This set spans both linear and sublinear budgets, as well as capped variants that prevent excessive sampling for large $m$.

These choices are motivated by computational constraints, rather than theoretical optimality. The results of this sweep (Tables~\ref{tab:sparsek-m3}--\ref{tab:sparsek-m13}) demonstrate that Sparse-$K$ exhibits stable performance across a broad range of feature counts, indicating that the method is easily generalizable. To be concise we only display the top-3 best performing configurations per feature size.  

\begin{table}[h!]
\caption{Sparse-$K$ sweep results for feature size $M=(2\text{-}3)$, $N=96$. Values reported as mean $\pm$ standard deviation.}
\label{tab:sparsek-m3}
\renewcommand{\arraystretch}{1.25}
\begin{center}
\begin{tabular}{ccccc}
\hline
Rank & $(k,c)$ & IoU & F1 & AUROC \\
\hline
1 & $(4\sqrt{M},\,0.5M)$ & $0.903 \pm 0.237$ & $0.927 \pm 0.177$ & $0.891 \pm 0.266$ \\
2 & $(2\sqrt{M},\,0.5M)$ & $0.896 \pm 0.243$ & $0.922 \pm 0.182$ & $0.883 \pm 0.274$ \\
3 & $(\sqrt{M},\,0.5M)$  & $0.896 \pm 0.243$ & $0.922 \pm 0.182$ & $0.883 \pm 0.274$ \\
\hline
\end{tabular}
\end{center}
\end{table}

\begin{table}[h!]
\caption{Sparse-$K$ sweep results for feature size $M=(4\text{-}5)$, $N=100$. Values reported as mean $\pm$ standard deviation.}
\label{tab:sparsek-m5}
\renewcommand{\arraystretch}{1.25}
\begin{center}
\begin{tabular}{ccccc}
\hline
Rank & $(k,c)$ & IoU & F1 & AUROC \\
\hline
1 & $(2\sqrt{M},\,M)$            & $0.561 \pm 0.351$ & $0.652 \pm 0.307$ & $0.653 \pm 0.297$ \\
2 & $(2\sqrt{M},\,\min(M,16k))$  & $0.554 \pm 0.341$ & $0.650 \pm 0.295$ & $0.651 \pm 0.285$ \\
3 & $(2\sqrt{M},\,\min(M,4k))$   & $0.544 \pm 0.329$ & $0.648 \pm 0.274$ & $0.651 \pm 0.269$ \\
\hline
\end{tabular}
\end{center}
\end{table}

\begin{table}[h!]
\caption{Sparse-$K$ sweep results for feature size $M=(6\text{-}8)$, $N=99$. Values reported as mean $\pm$ standard deviation.}
\label{tab:sparsek-m8}
\renewcommand{\arraystretch}{1.25}
\begin{center}
\begin{tabular}{ccccc}
\hline
Rank & $(k,c)$ & IoU & F1 & AUROC \\
\hline
1 & $(4\sqrt{M},\,\min(M,8k))$ & $0.445 \pm 0.330$ & $0.546 \pm 0.318$ & $0.641 \pm 0.253$ \\
2 & $(\sqrt{M},\,\min(M,4k))$  & $0.424 \pm 0.305$ & $0.535 \pm 0.295$ & $0.634 \pm 0.235$ \\
3 & $(4\sqrt{M},\,M)$          & $0.431 \pm 0.336$ & $0.529 \pm 0.326$ & $0.625 \pm 0.262$ \\
\hline
\end{tabular}
\end{center}
\end{table}

\begin{table}[h!]
\caption{Sparse-$K$ sweep results for feature size $M=(9\text{-}11)$, $N=99$. Values reported as mean $\pm$ standard deviation.}
\label{tab:sparsek-m11}
\renewcommand{\arraystretch}{1.25}
\begin{center}
\begin{tabular}{ccccc}
\hline
Rank & $(k,c)$ & IoU & F1 & AUROC \\
\hline
1 & $(4\sqrt{M},\,M)$           & $0.353 \pm 0.306$ & $0.455 \pm 0.306$ & $0.623 \pm 0.207$ \\
2 & $(2\sqrt{M},\,\min(M,16k))$ & $0.355 \pm 0.310$ & $0.455 \pm 0.312$ & $0.622 \pm 0.213$ \\
3 & $(4\sqrt{M},\,\min(M,16k))$ & $0.342 \pm 0.292$ & $0.448 \pm 0.298$ & $0.619 \pm 0.200$ \\
\hline
\end{tabular}
\end{center}
\end{table}

\begin{table}[h!]
\caption{Sparse-$K$ sweep results for feature size $M=(12+)$, $N=24$. Values reported as mean $\pm$ standard deviation.}
\label{tab:sparsek-m13}
\renewcommand{\arraystretch}{1.25}
\begin{center}
\begin{tabular}{ccccc}
\hline
Rank & $(k,c)$ & IoU & F1 & AUROC \\
\hline
1 & $(4\sqrt{M},\,\min(M,4k))$ & $0.632 \pm 0.393$ & $0.694 \pm 0.353$ & $0.815 \pm 0.212$ \\
2 & $(2\sqrt{M},\,\min(M,4k))$ & $0.421 \pm 0.418$ & $0.476 \pm 0.421$ & $0.683 \pm 0.251$ \\
3 & $(2\sqrt{M},\,0.25M)$      & $0.323 \pm 0.351$ & $0.399 \pm 0.359$ & $0.634 \pm 0.216$ \\
\hline
\end{tabular}
\end{center}
\end{table}

\end{document}